\documentclass[runningheads]{llncs}
\usepackage{graphicx}

\usepackage{tikz}
\usepackage{comment}
\usepackage{amsmath,amssymb} 
\usepackage{color}
\usepackage{wrapfig}
\usepackage[export]{adjustbox}
\usepackage{caption}
\usepackage{booktabs}
\usepackage{algorithm}
\usepackage{algorithmic}
 \usepackage{array,multirow,graphicx}
 \usepackage{float}
\usepackage{bbold}
\usepackage{hyperref}
\hypersetup{pagebackref,breaklinks,colorlinks,bookmarks=false}
\usepackage[accsupp]{axessibility}
\begin{document}
\pagestyle{headings}
\mainmatter
\def\ECCVSubNumber{100}  

\title{PLMCL: Partial-Label Momentum Curriculum Learning for Multi-Label Image Classification} 

\titlerunning{PLMCL for Multi-Label Image Classification}
%
\author{Rabab Abdelfattah \inst{1} \and
Xin Zhang\inst{1} \and
Zhenyao Wu\inst{2} \and
Xinyi Wu\inst{2}\\
Xiaofeng Wang \inst{1}\and
Song Wang \inst{2}}
\authorrunning{Rabab Abdelfattah et al.}
%
\institute{Department of Electrical Engineering, University of South Carolina, USA \email{} \and
Department of Computer Science, University of South Carolina, USA \\ \email{\{rabab,xz8,zhenyao,xinyiw\}@email.sc.edu}, \{wangxi, songwang\}@cec.sc.edu}

\maketitle
\begin{abstract}
Multi-label image classification aims to predict all possible labels in an image. It is usually formulated as a partial-label learning problem, given the fact that it could be expensive in practice to annotate all labels in every training image. Existing works on partial-label learning focus on the case where each training image is annotated with only a subset of its labels.  A special case is to annotate only one positive label in each training image.
To further relieve the annotation burden and enhance the performance of the classifier, this paper proposes a new partial-label setting in which only a subset of the training images are labeled, each with only one positive label, while the rest of the training images remain unlabeled.
To handle this new setting, we propose an end-to-end deep network, PLMCL (Partial-Label Momentum Curriculum Learning), that can learn to produce confident pseudo labels for both partially-labeled and unlabeled training images.  The novel momentum-based law updates soft pseudo labels on each training image with the consideration of the updating velocity of pseudo labels, which help avoid trapping to low-confidence local minimum, especially at the early stage of training in lack of both observed labels and confidence on pseudo labels. In addition, we present a confidence-aware scheduler to adaptively perform easy-to-hard learning for different labels.
Extensive experiments demonstrate that our proposed PLMCL outperforms many state-of-the-art multi-label classification methods under various partial-label settings on three different datasets.
\end{abstract}

\section{Introduction}
\begin{figure*}[!t]
  \centering
    \includegraphics[scale=0.35]{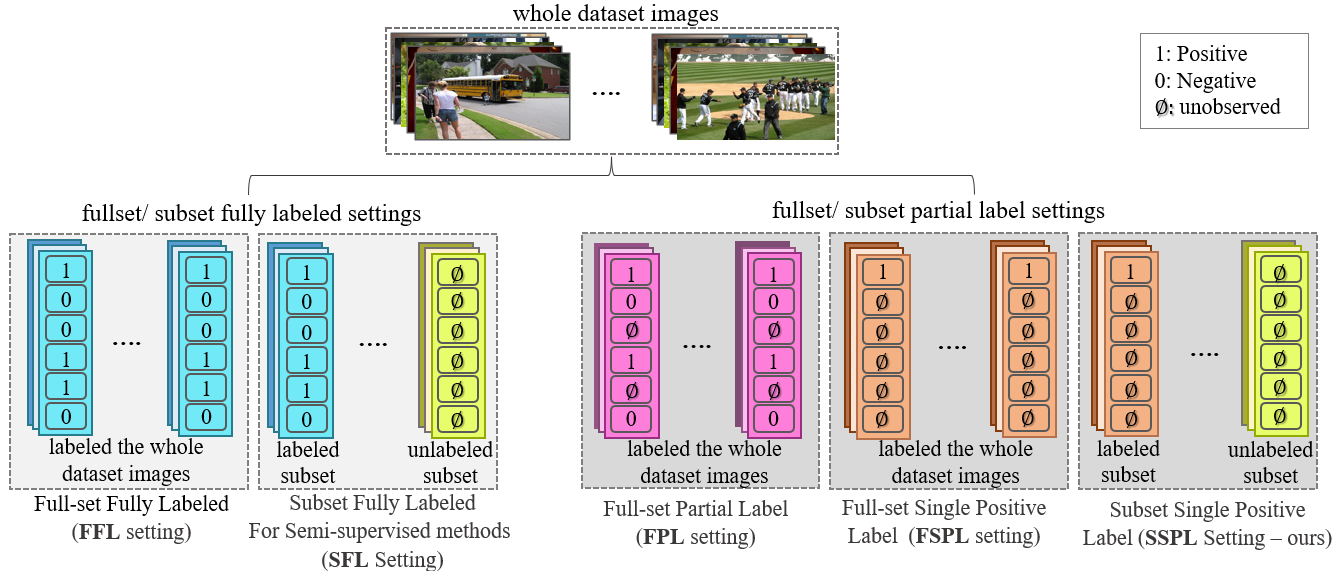}
  \vskip -3mm
  \caption{An illustration of various label settings.  \textbf{Full-set fully labeled (FFL)}: All positive and negative labels are annotated on each training image. \textbf{Full-set partial labels (FPL)}~\cite{durand2019learning}: At least one positive or negative label is annotated on each training image. \textbf{Full-set single positive labels (FSPL)}~\cite{cole2021multi}: Exactly one positive label is observed on each training image. \textbf{Subset fully labeled (SFL)}: All positive and negative labels are annotated on a subset of training images and the rest of the images are unlabeled. \textbf{Subset single positive labels (SSPL-ours)}: Exactly one positive label is annotated on each image from a subset of the training dataset while the rest of the images are unlabeled. 
}

\label{fig:label-type}
\vskip -6mm
\end{figure*}
With the advances in deep learning, significant progress has been made on single-label image classification problems~\cite{deng2009imagenet} where each image only has one label. However, in many real applications, one image may contain multiple objects and/or exhibit multiple attributes that cannot be well described by a single label. For instance, a scene may contain multiple objects and a CT scan may indicate multiple conditions.  This leads to an important computer-vision task of multi-label image classification that aims to identify all the labels on an image. A great challenge in multi-label image classification comes from the need of a large number of labeled training images. In particular, many supervised-learning algorithms require all labels on every training image to be accurately annotated, which can be very difficult and laborious~\cite{deng2014scalable}. 

To relieve the annotation burden of fully labeling, recent works on multi-label classification consider training the network with partial labels~\cite{cole2021multi,durand2019learning,huynh2020interactive,kundu2020exploiting,xu2013speedup,PinedaSalvador2019im2set}. 
One typical partial-label setting is \textbf{full-set partial labels (FPL)}~\cite{durand2019learning}, which annotates only a subset of labels on each training image.  
Following this setting, a special case is \textbf{full-set single positive labels (FSPL)}~\cite{cole2021multi}, which annotates only one positive label on each image (Fig.~\ref{fig:label-type}). 
Although these partial-label settings can, to some extent, mitigate the annotation burden, we still need to go through all training images for annotation.

To go one step further, we consider a new partial-label setting of \textbf{subset single positive labels (SSPL)}, where the training dataset consists of both labeled and unlabeled images, while, for the subset of labeled images, only one positive label is annotated on each image.
Clearly, this new setting can further reduce the annotation cost for large-scale multi-label datasets and practically our setting can be utilized by only annotating a limited percentage of total images with one single positive label per image.  For
instance, COCO dataset has 2.5M labels which takes 20K worker hours to identify categories~\cite{lin2014microsoft}, while 60\% SSPL only requires 49.2k labels and the annotation time is up to 394 hours. Here we did not even account for the fact that detecting absence can be more difficult than detecting presence.
However, the inclusion of unlabeled images significantly increases the difficulty of the network design and training. 
In particular, those existing partial-label learning methods based on label correlation~\cite{huynh2020interactive} and label matrix completion~\cite{cabral2011matrix} are not applicable to our new setting since they cannot handle single positive labels and unlabeled data~simultaneously.  From another point of view, although our proposed setting and semi-supervised learning both use unlabeled subsets, the performance of semi-supervised models may degrade when training on the SSPL setting. This is because semi-supervised methods usually assume access to a subset of fully-labeled images (SFL setting) to initialize the training process, while such a fully-labeled subset is not available on the SSPL setting as shown in Fig.~\ref{fig:label-type}. Such degradation can be observed in the comparison experiments presented in Section~\ref{sub:ablation}.

This paper presents a new Partial-Label Momentum Curriculum Learning (PLMCL) method for end-to-end training of the multi-label classifier under the proposed new partial-label setting. 
In particular, a set of soft pseudo labels for each image are estimated and dynamically updated by momentum. The momentum combines the classifier predictions and the pseudo labels to identify the direction for updating the pseudo labels.  
Meanwhile, to update the multi-label classifier parameters, we introduce a scheduled loss function based on standard cross-entropy with which the learning gradually moves from observed (easy) labels to unobserved (hard) labels. Our proposed PLMCL can also be easily extended to handle other partial-label settings~\cite{cole2021multi,durand2019learning,huynh2020interactive} for multi-label image classification. 

Our contributions are summarized as follows:
\begin{itemize}
\item
We introduce a new partial-label setting, SSPL, for multi-label image classification by allowing the inclusion of unlabeled images as training images. This setting can be as simple as the case where only a subset of images have labels and only one single positive label is annotated per image in this subset.  This new setting can significantly reduce the annotation burden as well as leverage more unlabeled data for multi-label image classification.
\item 
We propose a new PLMCL framework for multi-label image classification under different partial-label settings. The major novelty comes from the introduction of the momentum and the confidence levels into the partial-label setting.  Instead of accelerating learning as discussed in~\cite{goodfellow2016deep}, the momentum in this paper actually brings a stronger stability requirement into training, which focuses on the convergence of not only the pseudo labels, but also the updating velocity of pseudo labels that indicates high confidence on the pseudo labels.  This is important, especially at the early stage of training when the observed labels are lack and the confidence levels on the pseudo labels are low. Otherwise, premature convergence could be achieved with low confidence levels due to lack of observed labels.  
\item
Extensive experimental results show that our method outperforms the state-of-the-arts under three different partial-label settings on three widely-used datasets. With fewer observed labels in our proposed setting, our method can still get comparable classification results as those methods using full-set single positive labels (FSPL) setting, as shown in Fig.~\ref{fig:label-type}.
\end{itemize}
The rest of the paper is organized as follows. Section~\ref{sec:rw} briefly reviews the related works. Section~\ref{sec:method} presents the details of the proposed method. The experiment results are demonstrated in Section~\ref{sec:experiments}. Finally, the conclusions are drawn in Section~\ref{sec:con}.

\section{Related Work}
\label{sec:rw}

This section briefly reviews the related work on partial-label learning under different settings (Fig.~\ref{fig:label-type}) as well as semi-supervised learning and curriculum learning methods.
\subsection{Partial-label Learning under Different Settings}

\noindent
\textbf{Full-set Partial Labels (FPL).}
\label{subsub:labels} Early work on multi-label image classification under the FPL setting assumes that the unobserved labels are negatives~\cite{sun2017revisiting,mahajan2018exploring,bucak2011multi,chen2013fast}. This assumption inevitably introduces false negative labels during the training and therefore will lead to a significant performance degradation in classification. Another direction is to estimate the unobserved labels based on the correlation between the labels~\cite{mahajan2018exploring,wu2015ml}, label matrix completion~\cite{wang2014binary,cabral2011matrix}, low-rank learning~\cite{yu2014large,yang2016improving}, and probabilistic models~\cite{kapoor2012multilabel,chu2018deep}. Most of these models, however, have to solve an optimization problem during the training, which might be computationally expensive for deep learning given the high complexity of deep neural networks~\cite{durand2019learning}. Some recent work uses end-to-end deep learning models to predict unobserved labels. For instance, the method in~\cite{huynh2020interactive} takes advantage of image/label similarity graphs and builds the dependencies among the labels based on the label co-occurrence information. Nevertheless, it requires more than one label per image to build these relations and therefore is not applicable for single-label settings. Another work in~\cite{durand2019learning} integrates graph neural network with curriculum learning. The proposed strategy predict the unobserved labels based on network models and add those predicted labels into the training dataset using a threshold-based strategy.  Those selected labels are called ``weakly labels'', which may change over epochs. Therefore, there is a possibility that the predictions of some unobserved labels are never selected into the training, which may result in information loss.  In contrast, our model learns the pseudo labels dynamically based on the momentum throughout the training process. Different from~~\cite{durand2019learning}, every pseudo label will more or less contribute to the training in each epoch, depending on its confidence level.  This can guarantee the continuity in pseudo label updates, which is lack from the threshold-based strategy.

\smallskip
\noindent
\textbf{Full-set Single Positive Label (FSPL).} This setting is introduced in~\cite{cole2021multi}, where the authors present an end-to-end model that jointly trains the image classifier and the label estimator for online estimation of unobserved labels. The label estimator is implemented to build a simple look-up table which requires to store a matrix containing the whole dataset images, including all the previously estimated labels per image, in memory and update it in each epoch during the entire training process.
Obviously, this method requires ample memory space to store the full label matrix for all training images, which makes it almost infeasible for large datasets or large numbers of labels. Moreover, this method relies on random initialization for the label estimator and assigns equal weights for the observed and unobserved terms in the suggested loss function, which may lead to degradation in the classifier's performance. In contrast, our method initializes the pseudo labels with unbiased probability values guaranteeing stable training using a self-guided momentum factor. Therefore, the self-guided factor works to (i) identify the amount of change should the labels get to move towards 1 or 0 starting from unbiased probability, and (ii) reduce the amount of change for the labels with a high confidence score. In addition, our momentum-based updating method does not require keeping the full labels in the memory since we only keep $B$ vectors in memory per each batch, where $B$ is the batch size. Finally, our method uses a scheduled loss function to give different weights for the unobserved term, which can gradually move the learning from the observed labels to the unobserved labels.

 \smallskip
\noindent
\textbf{Subset Single Positive Labels (SSPL)}. This is our proposed setting which assumes that each labeled image is annotated with only a single positive label, while the unlabeled images are totally unlabeled.  A practical scenario is iNaturalist dataset, where each image only has one positive label while the other classes are unlabeled ~\cite{cole2021multi}.  Adding new unlabeled images to enrich this dataset directly leads to the SSPL setting.  Actually, this applies to most datasets for multi-class classification.  Adding new unlabeled images makes those datasets suitable for the multi-label problem under the SSPL setting without annotation costs. To the best of our knowledge, there are no approaches specifically designed for SSPL.  However, some semi-supervised models can be modified to fit SSPL setting.  We will go through these methods in the next subsection.
Although these methods ~\cite{kundu2020exploiting,mac2019presence,cole2021multi} can be easily applied to the SSPL setting, these methods may still not handle the unlabeled set of the SSPL properly. Since these models usually assign assumptions for the unobserved labels, such as ignoring the unobserved labels, assuming negative, giving them a down-weight, or starting with a random probability. Therefore,  the whole labels per image, in unlabeled set, are addressed based on any of the previously mentioned assumptions during the training. Consequently, the SSPL unlabeled set's images negatively impact the performance of these models. 
\subsection{Semi-supervised Learning}
\label{semi-supervised}
Most of semi-supervised learning (SSL) methods on multi-class classification, while a few of them study multi-label classification~\cite{arazo2020pseudo,berthelot2019remixmatch,chapelle2009semi,liu2006semi,laine2016temporal,niu2019multi,rizve2021in,sohn2020fixmatch,tarvainen2017mean,wang2018adaptive,wang2013dynamic}.  Among those methods, it is usually assumed that some of the training images are fully labeled while the others are totally unlabeled, i.e., the SFL setting in Fig. \ref{fig:label-type}. Most state-of-the-art methods create pseudo labels for the unlabeled data based on different approaches such as self-training-based and consistency-based approaches. Self-training-based approaches follow these steps to train the semi-supervised model: (i) train the  model based on the labeled data, (ii) use the trained model to get the predictions for the unlabeled data, and (iii) apply the threshold-based strategy to the predictions to select the pseudo labels for unobserved data, which are considered only for the predictions greater than the threshold value i.e.,~\cite{arazo2020pseudo,rizve2021in}.  Therefore, the selected pseudo labels are used in the loss function for training in the next epoch.   
On the other hand, the consistency-based approaches follow this strategy: (i) find the supervised loss based on the labeled data, (ii) apply the data perturbations, i.e., data augmentation and stochastic regularization, to produce two versions for the same image, and (iii) use the prediction of one image-version after applying the threshold as the pseudo label for the other image-version~\cite{berthelot2019remixmatch,sohn2020fixmatch,tarvainen2017mean}.
To be concluded, most of semi-supervised models generate the pseudo labels based on the predictions that greater than the threshold value.
To make the semi-supervised models applicable for SSPL setting, the unobserved labels in the labeled set of SSPL are ignored since the labeled data of SSPL only contains one single label per image. Therefore, although the semi-supervised can be directly trained on SSPL setting, the performance of these models significantly decreased since the supervised loss function only works with a single label as discussed in subsection~\ref{sub:ablation}. While in our case, all the unobserved labels, in labeled and unlabeled sets of SSPL, initiate with unbiased probabilities to consider as initial pseudo labels and then improve their probabilities during the training using our momentum-based method. We also adapt some semi-supervised models to be applicable for multi-label classification problems, by replacing Softmax layer with Sigmoid, that are designed for multi-class classification, such as~\cite{sohn2020fixmatch}. 
\subsection{Curriculum Learning}
Curriculum learning~\cite{bengio2009curriculum} has been applied to many applications related to multi-labels or multi-classes. Given the training data with fully observed labels, different iterative self-paced learning algorithms~\cite{kumar2010self,jiang2015self} have been developed to enhance traditional image classification, object localization, and multimedia event detection -- in each iteration, easy samples are selected and the model parameters are updated accordingly. 
Recently, CurriculumNet~\cite{guo2018curriculumnet} is proposed to learn knowledge from large-scale noisy web images by following the curriculum learning strategy. However, this strategy employs a clustering-based model to assess the complexity of the image and cannot be used for multi-label image classification~\cite{durand2019learning}. Our PLMCL also follows the general idea of curriculum learning~\cite{wang2021survey} by
starting the learning on observed labels, and then progressively moving to the unobserved labels.

\smallskip
\noindent
\section{The PLMCL Method}
\label{sec:method}
Assume that the classification network is trained on the image dataset $\mathcal I$ for multi-label classification. Since we focus on FSPL and SSPL settings, let $\mathcal D$ be the subset of $\mathcal I$, in which each image has only one positive label and the images in $\mathcal I \backslash \mathcal D$ are totally unlabeled. Note that $\mathcal D\subset \mathcal I$ means SSPL setting and $\mathcal D = \mathcal I$ means FSPL setting. Let $L$ be the total number of classes. For each image $i \in \mathcal D$, we denote $y_o^i\in \{\emptyset,1\}^L$ as the observed label vector where each entry of $y_o^i$ can be 1 (observed positive) or $\emptyset$ (unlabeled). Meantime, since the network is trained over epochs, we use $\hat{y}^i_{u,t} \in [0,1]^L$ to denote the soft pseudo label vector of of image $i$ at epoch $t$~\cite{tanaka2018joint}. Let $f$ denote the classification network and $\theta_t$ be the network parameter obtained at epoch $t$.  Given an image $i\in \mathcal I$, the predicted labels from the network is defined as $\hat{y}^i_t = f(i, \theta_t) \in [0,1]^L $. The binary cross-entropy loss function between two scalars $p,q \in [0,1]$ is defined as
$\mathcal{L}(p,q)=-p\log(q)-(1-p)\log(1-q)$. 
Given $v\in \mathbb R^L$, we use $[v]_j$ to denote the $j$th entry of $v$. For notation convenience, we drop the index $t$ if it is clear in context.
\begin{figure*}[!t]
  \centering
  			\includegraphics[scale=0.365]{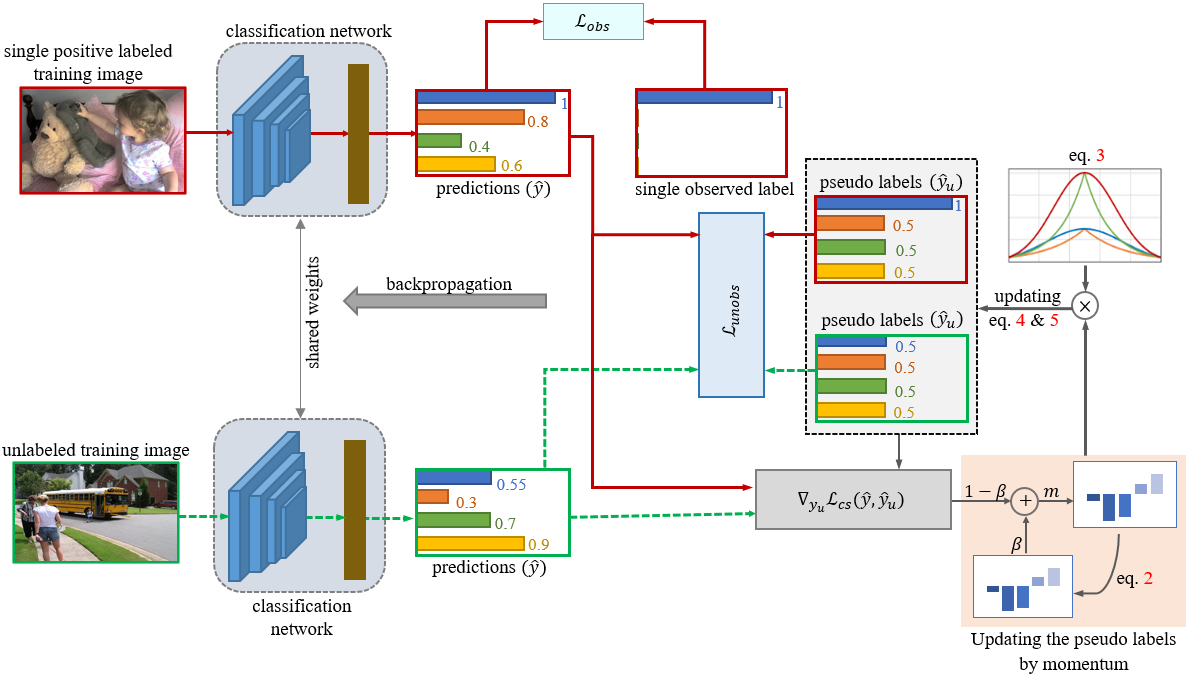}
  			    	\vskip-3mm
  \caption{An overview of the PLMCL method that consists of classification network, backbone and one-layer classifier, to generate a prediction for an input image. The classification network is end-to-end trained using the scheduled cross-entropy loss function. The pseudo labels $\hat y_{u,t}$ are updated based on the momentum $m_t$, which is the weighted sum of the last momentum $m_{t-1}$ and the gradient of the loss between the predicted labels $\hat y_{t-1}$ and the pseudo labels $\hat y_{u,t-1}$ (Orange region). At the test, only the classification network is applied.}
      	\vskip-4.9mm
  \label{fig:PLMCL}
\end{figure*}
\subsection{Overview}
The PLMCL method is proposed for multi-label classification to deal with the case where each image in $\mathcal D$ only contains a single positive label. The procedure of PLMCL is shown in Fig.~\ref{fig:PLMCL}. During training, the pseudo label vector $\hat{y}_{u,0}$ of each image in~$\mathcal I$ will be initialized with the same values of $y_o$ for the observed labels and the unbiased probability $0.5$ for the unobserved labels. Then PLMCL updates the pseudo label $\hat y_{u,t}$ based on the momentum $m_t$, which is the weighted sum of the last momentum $m_{t-1}$ and the gradient of the loss between the predicted label vector $\hat y_{t-1}$ and the pseudo label vector $\hat y_{u,t-1}$ (Orange region in Fig.~\ref{fig:PLMCL}).  It is expected that the pseudo labels are adaptively refined to provide stability for the confident labels and high momentum for the unconfident labels.  The classification network is only employed during the test to predict the labels.
 
\subsection{Updating Pseudo Labels by Momentum}

\noindent
\textbf{Momentum Generation.} When $\mathcal D$, the subset of the labeled images, is small, training has to start with a very limited number of observed labels, and therefore the resulting predictions will be inaccurate.  In this case we cannot directly aggregate the predictions from the previous epoch to the current epoch using the off-the-shelf temporal ensembling techniques~\cite{laine2016temporal} to update the pseudo labels. Otherwise, the aggregation may soon arrive at a local minimum with low prediction confidence.  Instead, our idea is to exploit the aggregation of the gradient (the momentum~\cite{goodfellow2016deep}) to update the pseudo label vector (Section~\ref{sub:ablation} includes ablation study to compare these two approaches). By this way, we can help avoid undesired local minima~\cite{sariyildiz2019gradient}.  

To do so, we consider the binary cross-entropy loss function
\begin{align}
\mathcal L_{cs}(\hat y_{t-1},\hat{y}_{u,t-1}) = \frac{1}{L}\sum_{j=1}^L \mathcal L\left([\hat y_{t-1}]_j,[\hat{y}_{u,t-1}]_j\right)
\end{align}
and introduce the momentum vector $m_t$, which is updated based on the gradient of the loss function at epoch $t-1$, i.e., $\nabla \mathcal{L}_{cs} (\hat y_{t-1},\hat{y}_{u,t-1})$:   
\begin{equation}
m_t = \beta_1 m_{t-1} + (1-\beta_1) \nabla \mathcal{L}_{cs} (\hat y_{t-1},\hat{y}_{u,t-1}),
\label{eq:momentum}
\end{equation}
where $\beta_1$ is moving-average decay scalar and $m_{0}$ is initialized by zero. Notice that even when $\nabla \mathcal{L}_{cs} =0$ (which means that the loss reaches its local minima), the momentum will still be updated, which can therefore avoid premature convergence.  This is especially important at the early stage of training.  Another observation is that, when one entry of the momentum vector consistently moves towards the same direction, the estimation of the associated pseudo label will be more confident and the variations can be well smoothed over consequence epochs which leads to a more stable training behavior.

\smallskip
\noindent
\textbf{Self-Guided Momentum Factor.} Intuitively, when the confidence of the pseudo label for a class is low, we want the corresponding momentum to contribute more to the pseudo label's iteration, in order to drive the pseudo label away from its current value to avoid undesired local minima. Otherwise, we want to keep the pseudo label, in which case the momentum should contribute less.  With this idea, we introduce the self-guided momentum factor $\psi(\hat y_{u,t}) \in \mathbb R^L$ at epoch $t$:
\begin{equation}
    \psi (\hat{y}_{u,t}) = \alpha \exp ( - \lambda |2\hat{y}_{u,t}-{\bf 1}|^n),
    \label{eq:psi}
\end{equation}
where $\alpha > 0 $ is the curriculum learning rate, $\lambda > 0 $ and $n$ are tuned parameters, and ${\bf 1} \in \mathbb R^L$ is the vector with 1 at each entry. The exponential function $\exp$ and the absolute value $|\cdot|$ are operated element-wisely.  In all our experiments, we assign $n$ with order 2.
Notice that the confidence of a pseudo label is reflected by the term $|2\hat{y}_{u,t}-{\bf 1}|$ in the definition of $\psi$. For instance, if $[\hat{y}_{u,t}]_j$, the $j$th entry of $\hat{y}_{u,t}$, is 0.5, the related momentum factor will achieve its maximal value $\alpha$, which means that our module is not confident about the pseudo label and the related momentum should contribute more in the next iteration of the pseudo label.  Otherwise, if $[\hat{y}_{u,t}]_j = 0$ or $[\hat{y}_{u,t}]_j = 1$, 
the related momentum factor reaches its minimum $\alpha\exp(-\lambda)$ (because $[\hat{y}_{u,t}]_j \in [0,1]$), which indicates high confidence on the current pseudo label and the momentum should contribute less so that the value of the pseudo label can be maximally kept. 

With the self-guided momentum factor, we can update the pseudo label latent parameters  $y_{u,t} \in \mathbb R^L$ by:
\begin{equation}
y_{u,t}=y_{u,t-1} - \psi (\hat{y}_{u,t-1})\circ  m_t
\label{eq:update_vector}
\end{equation} 
where $\circ$ means the element-wise multiplication. From this equation, we can see that the momentum $m_t$ actually indicates a weighted difference between $y_{u,t}$ and $y_{u,t-1}$.  Keeping this in mind, Equation~\eqref{eq:momentum} yields an interesting observation: Our approach expects not only the convergence of the pseudo labels (as it is in~\cite{laine2016temporal}), but also the convergence of the updating velocity of the pseudo labels. By doing so, unexpected local minima can be effectively avoided~\cite{goodfellow2016deep}.  

Notice that the entry of $y_{u,t}$ might be outside the interval $[0,1]$.  So we need to regulate $y_{u,t}$ to obtain $\hat y_{u,t}$ by
\begin{equation}
[\hat{y}_{u,t}]_j = \sigma([y_{u,t}]_j)
\end{equation}
for $j=1,2,\cdots,L$, where $\sigma:\mathbb{R} \to [0, 1]$ is the sigmoid function that maps each entry of $y_{u,t}$ into $[0,1]$. In practice, the gradient in equation~\eqref{eq:momentum} is actually with respect to ${y}_{u,t-1}$, i.e., $\nabla \mathcal{L}_{cs}  = \nabla_{{y}_{u,t-1}} \mathcal{L}_{cs} (\hat y_{t-1},\hat{y}_{u,t-1})$.
\subsection{Scheduled Loss Function}
\label{sub:loss-function}
To learn the parameters of the primary multi-label CNN, the following loss function must be minimized
\begin{equation}
    \mathcal{L}_{PLMCL} = \mathcal{L}_{obs} + \mathcal{L}_{unobs},
    \label{eq:loss}
\end{equation}
where $\mathcal{L}_{obs}$ is the cross-entropy loss between the predicted labels and the ground truth from the observed labels plus the regularizer~\cite{cole2021multi} and  $\mathcal{L}_{unobs}$ is the weighted cross-entropy loss between the predicted labels and the pseudo labels for the unobserved labels 
\begin{align}
    \mathcal{L}_{unobs} = \sum_{i \in \mathcal I} \sum_{j\in \mathcal U_i} \xi([\hat y_{u,t}]_j,\varphi_t) \mathcal L\left([\hat y_{t}]_j,[\hat{y}_{u,t}]_j\right).
\end{align} 
Here $\mathcal U_i$ is the set of the unobserved labels on image~$i$ and $\xi([\hat y_{u,t}]_j,\varphi_t)$ is a new scheduler that weights the unobserved loss, defined by
\begin{align}
    \xi([\hat y_{u,t}]_j,\varphi_t) =\beta_2 \frac{1-\gamma \exp(-10 |2[\hat{y}_{u,t}]_j-1|) }{1+ \gamma \exp(-10|2[\hat{y}_{u,t}]_j-1|) },
    \label{eq:sched}
\end{align}
where $\beta_2$ is the positive hyper-parameter,   $\gamma = 1- \varphi_t$, and $\varphi_t$ is the current epoch $t$ divided by the total number of epochs.
This scheduler plays a critical role in balancing the contributions of unobserved labels to the total loss and the potential negative impacts of false pseudo labels.  When a pseudo label has low confidence level, the related loss is in appropriate, the scheduler will assign a small weight to this loss such that the total loss $\mathcal{L}_{unobs}$ will not be significantly affected. This is especially important at the beginning of the training process when the pseudo labels are still not indicative.
Different from most previous schedulers relying on the training steps or amount of available training data~\cite{jean2019adaptive}, our scheduler is adaptive, counting for not only the training steps, but the validation performance during training as well.  To be specific, our scheduler $\xi$ uses the confidence $|2[\hat{y}_{u,t}]_j-1|$ to determine the scheduler value. When $[\hat{y}_{u,t}]_j$ is close to 0.5, $|2[\hat{y}_{u,t}]_j-1|$ will be close to 0 and therefore the value of $\xi$ will be small. Over epochs, the pseudo labels can gradually build up their confidence and start to converge.  In that case, the value of $\xi$ will increase.

We use soft pseudo labels, instead of sharp $1$ or $0$, in $\mathcal{L}_{unobs}$, because the latter may make the predictions of the output layer over-confident~\cite{szegedy2016rethinking}. In contrast, using the pseudo labels as the distribution scores helps smooth the predictions, which is consistent with the idea of label-smoothing regularization~\cite{szegedy2016rethinking}.
\begin{algorithm}[!t]
	\caption{Momentum Curriculum Algorithm}
	\label{alg:train}
	\begin{algorithmic}[1]
		\REQUIRE  input image $i \in \mathcal{I}$;
		\REQUIRE observed label $y_{o}$;
		\REQUIRE  neural network $f(i, \theta)$ with trainable parameters $\theta$ and input $i$.
		\REQUIRE 
        $[\hat{y}_{u}]_j = 1 \mbox{ or } 0$,~ if labeled\\
        $[\hat{y}_{u}]_j = 0.5$,~ if unlabeled\\
		\STATE \textbf{Repeat}
		\begin{ALC@g}
    		\STATE $\hat{y} \longleftarrow f(i, \theta)$;
             \STATE $m \longleftarrow \beta_1 m + (1-\beta_1) \nabla_{y_{u}} \mathcal{L}_{cs} (\hat y,\hat{y}_{u})$;
    	    \STATE $y_{u} \longleftarrow y_{u} -  \psi(\hat{y}_u)  m$;
    	    \STATE $\mathcal L_{PLMCL} \longleftarrow \mathcal L_{obs}+ \mathcal L_{unobs}$;
    	    \STATE update $\theta$ via backpropagation to improve the classification model based on curriculum framework;
    	    \STATE $\hat{y}_{u} \longleftarrow \sigma(y_{u})$;
	    \end{ALC@g}
        \STATE \textbf{until} the max iteration or convergence;
        \STATE \textbf{Output} optimal CNN parameters, $\hat{y}_u$ able to guide the prediction $\hat{y}$ to be very close from true label.
	\end{algorithmic}
\end{algorithm} 
\section{Experiments}
\label{sec:experiments}
\subsection{Datasets Preparation}
\textbf{Datasets.} The \textbf{PASCAL VOC} dataset~\cite{everingham2011pascal} contains natural images from 20 different classes, including 5,717 training images and 5,823 images in the official validation set for testing. The \textbf{MS-COCO} dataset~\cite{lin2014microsoft} consists of 80 classes, including 82,081 training images and 40,137 testing images. The \textbf{NUS-WIDE} dateset~\cite{chua2009nus} consists of nearly 150K color images with various resolutions for training and 60.2K for testing. This dataset is associated with 81 classes. 
\begin{wraptable}{r}{6.5cm}
\centering
\vskip -1.5mm
\caption{Quantitative results (mAP) of multi-label image classification on different subsets of the single observed label (SSPL) setting on COCO, VOC, and NUS datasets. Bold represents the highest mAP and underline represents the second-best.}
\vskip-4mm
    \begin{adjustbox}{width=0.53\textwidth}  
    \begin{tabular}{|l|l|l|l|l||l|l|l|l|}
    \hline
    &\multicolumn{8}{c|}{\textbf{COCO dataset}}\\
\cline{2-9}
Losses&\multicolumn{4}{c||}{end-to-end}&\multicolumn{4}{c|}{linear-init}\\
\cline{2-9}
        ~ & 20\% & 40\% & 60\% & 80\% & 20\% & 40\% & 60\% & 80\%  \\ \hline
        $\mathcal{L}_{AN}$ & 46.3 & 53.8 & 59.5 & 62.4 & 53.4 & 61.1 & 63.8 & 65.6  \\ \hline
        $\mathcal{L}_{AN-LS}$ & 48.9 & 57.9 & 62.3 & \underline{65.5} & 58.2 & 62.8 & \underline{67.6} & 68.5  \\ \hline
        $\mathcal{L}_{WAN}$ & \underline{57.0} & \underline{60.9} & \underline{63.5} & 64.5 & 60.2 & 62.4 & 64.1 & 66.1  \\ \hline
        $\mathcal{L}_{EPR}$ & 52.7 & 58.4 & 61.5 & 62.6 & 61.0 & 63.6 & 65.3 & 66.6  \\ \hline
        $\mathcal{L}_{ROLE}$ & 47.3 & 57.6 & 62.7 & 65.2 & \underline{61.9} & \underline{65.8} & 67.4 & \underline{68.6}  \\ \hline
        $\mathcal{L}_{PLMCL}$ & \textbf{61.0} & \textbf{65.1} & \textbf{68.4} & \textbf{70.4} & \textbf{64.8} & \textbf{67.8} & \textbf{69.2} & \textbf{71.5} \\ \hline
    \end{tabular}
    \end{adjustbox}
    \begin{adjustbox}{width=0.53\textwidth}  
    \begin{tabular}{|l|l|l|l|l||l|l|l|l|}
    \hline
        &\multicolumn{8}{c|}{\textbf{VOC dataset}}\\
\cline{2-9}
Losses&\multicolumn{4}{c||}{end-to-end}&\multicolumn{4}{c|}{linear-init}\\
\cline{2-9}
        ~ & 20\% & 40\% & 60\% & 80\% & 20\% & 40\% & 60\% & 80\%  \\ \hline
        $\mathcal{L}_{AN}$ & 51.3 & 71.7 & 80.2 & 82.8 & 78.6 & 81.4 & 83.7 & 84.9  \\ \hline
        $\mathcal{L}_{AN-LS}$ & 70.0 & 79.0 & 85.0 & 86.1 & 71.5 & 81.4 & 84.9 & 86.4  \\ \hline
        $\mathcal{L}_{WAN}$ & \underline{76.4} & \underline{82.5} & 85.1 & 85.6 & 79.4 & 83.6 & \underline{85.9} & 86.9  \\ \hline
        $\mathcal{L}_{EPR}$ & 75.5 & 81.0 & 83.9 & 84.5 & \underline{82.6} & \underline{84.9} & \underline{85.9} & 86.9  \\ \hline
        $\mathcal{L}_{ROLE}$ & 66.9 & 80.9 & \underline{85.9} & \underline{86.8} & 73.9 & 82.3 & 85.7 & \underline{87.5}  \\ \hline
        $\mathcal{L}_{PLMCL}$ & \textbf{79.9} & \textbf{84.6} & \textbf{87.6} & \textbf{88.4} & \textbf{83.1} & \textbf{85.4} & \textbf{87.8} & \textbf{87.7}  \\ \hline
    \end{tabular}
    \end{adjustbox}
    \begin{adjustbox}{width=0.53\textwidth}  
    \begin{tabular}{|l|l|l|l|l||l|l|l|l|}
    \hline
        &\multicolumn{8}{c|}{\textbf{NUS dataset}}\\
\cline{2-9}
Losses&\multicolumn{4}{c||}{end-to-end}&\multicolumn{4}{c|}{linear-init}\\
\cline{2-9}
        ~ & 20\% & 40\% & 60\% & 80\% & 20\% & 40\% & 60\% & 80\%  \\ \hline
        $\mathcal{L}_{AN}$ & 28.5 & 35.2 & 38.9 & 40.8 & 35.2 & 42.8 & 44.8 & 46.5  \\ \hline
        $\mathcal{L}_{AN-LS}$ & 27.8 & 36.0 & 39.1 & 41.4 & 39.1 & 44.4 & \underline{47.2} & 48.0  \\ \hline
        $\mathcal{L}_{WAN}$ & \underline{37.6} & \underline{41.3} & \underline{43.7} & \underline{44.8} & 41.4 & 44.5 & 45.7 & 46.2  \\ \hline
        $\mathcal{L}_{EPR}$ & 34.9 & 39.5 & 42.3 & 44.2 & \underline{43.4} & \underline{46.0} & \underline{47.2} & \underline{48.2}  \\ \hline
        $\mathcal{L}_{ROLE}$ & 27.0 & 32.2 & 37.1 & 39.7 & 36.3 & 42.8 & 46.4 & 47.5  \\ \hline
        $\mathcal{L}_{PLMCL}$ & \textbf{37.9} & \textbf{42.5} & \textbf{44.8} & \textbf{46.6} & \textbf{46.2} & \textbf{49.4} & \textbf{50.3} & \textbf{50.7}  \\ \hline
     \end{tabular}
             \end{adjustbox}
\label{tab:coco_sspl}
\vskip-10mm
\end{wraptable}
\smallskip
\noindent
\textbf{Data Preparation.} The SSPL setting annotates certain percent of the training images ($20\%$ to $80\%$) with only a single positive label and the rest of the images are totally unlabeled. 
The FSPL setting has $100\%$ of the training images annotated with only one single positive label.  To do so, we follow~\cite{cole2021multi} by assigning randomly only one positive label for each image in the training set.  \\

\noindent
\textbf{Implementation Details.} 
For a fair comparison, we use ResNet-50 
to initialize the weights of the backbone architecture for all models. The learning rate is $10^{-5}$ to train the VOC and COCO datasets and $10^{-4}$ to train the NUS dataset. The batch size is chosen as $8$ for the VOC dataset, and $16$ for the COCO and NUS datasets. Training is performed in two ways: end-to-end and linear-init.  End-to-end training simultaneously updates the parameters of the backbone and the classifier for 10 epochs, while linear-init training first fixes the parameters of the backbone and trains the classifier for 25 epochs, and then fine-tunes the parameters of the classifier and the backbone for 5 epochs~\cite{cole2021multi}. The best mAP on the validation set is reported for all experiments. It is worth mentioning that all experiments on our model share the same set of hyper-parameters. The mean average precision~(mAP) is applied to evaluate the performance of different approaches for multi-label classification in our paper, similar to \cite{cole2021multi,huynh2020interactive}.

\subsection{Subset Single Positive Label (SSPL)}
\label{subsub:SSPL}
This subsection compares the performance of our model $\mathcal{L}_{PLMCL}$ with other state-of-the-art approaches that address the unobserved labels in multi-label classification under SSPL settings.  The comparison results are listed in Table~\ref{tab:coco_sspl}. We observe that $\mathcal{L}_{AN}$~\cite{kundu2020exploiting} has the worst performance since it assumes that all the unobserved labels are negatives.  This assumption adds more noisy labels (false negatives) into the training process. To overcome this negative influence, $\mathcal{L}_{AN-LS}$ is introduced using label smoothing on $\mathcal{L}_{AN}$. Following~\cite{cole2021multi}, we use $\mathcal{L}_{AN-LS}$ as a baseline. To go one step further in mitigating the deteriorated effect of noisy labels, some models tend to down-weight the term related to negative labels in the loss function~$\mathcal{L}_{WAN}$~\cite{mac2019presence}. Consequently, $\mathcal{L}_{WAN}$ improves the mAPs on different datasets as compared to $\mathcal{L}_{AN}$ and $\mathcal{L}_{AN-LS}$. Finally, expected positive regularization $\mathcal{L}_{EPR}$~\cite{cole2021multi} and online estimation of unobserved labels~$\mathcal{L}_{ROLE}$ are introduced in \cite{cole2021multi} to deal with the single positive label, which is considered an extreme case of unobserved labels, in multi-label classification. 
We notice that $\mathcal{L}_{EPR}$~\cite{cole2021multi} provides stable performance  in different settings since $\mathcal{L}_{EPR}$ ignores the unobserved terms and focuses on finding the positive labels according to its regularizer.
Also notice that the performance of $\mathcal{L}_{ROLE}$ degrades a lot under 20\% SSPL in end-to-end setting since its classifier and label estimator are randomly initialized \cite{cole2021multi}. Compared with these methods, our proposed $\mathcal{L}_{PLMCL}$, based on the momentum, is superior in the three different datasets.
\begin{table*}[!t] \small
\caption{Quantitative results (mAP) of multi-label image classification for FSPL setting on three different datasets with end-to-end and linear-init training. Bold represents the highest mAP and underline represents the second-best.}
    \centering
    \begin{adjustbox}{width=1.01\textwidth}
    \begin{tabular}
        {
       | p{0.20\textwidth}
      |p{0.13\textwidth} ||
      >{\centering}p{0.13\textwidth} |
      >{\centering}p{0.13\textwidth} ||
      >{\centering}p{0.13\textwidth} |
      >{\centering}p{0.13\textwidth} ||
      >{\centering}p{0.13\textwidth} |
      >{\centering\arraybackslash}p{0.13\textwidth}|
    }
    \hline
        &~ &\multicolumn{2}{c||}{\textbf{COCO dataset}} &\multicolumn{2}{c||}{\textbf{VOC dataset}} &\multicolumn{2}{c|}{\textbf{NUS dataset}}   \\ \cline{3-8}
        Supervised by& Losses & end-to-end & linear-init & end-to-end & linear-init & end-to-end & linear-init  \\ \hline
       Fully Labeled &$\mathcal{L}_{BCE}$  & 75.5 & 76.8 & 89.1 & 90.1 & 52.6 & 54.7  \\ 
        &$\mathcal{L}_{BCE-LS}$ & 76.8 & 78.8 & 90 & 90.9 & 53.5 & 54.2  \\ \hline
        &$\mathcal{L}_{AN}$ & 64.1 & 67.3 & 85.1 & 87.7 & 42 & 46.8  \\ 
       &$\mathcal{L}_{AN-LS}$ & \underline{66.9} & \underline{69.2} & 86.7 & 86.5 & 44.9 & 50.5  \\ 
       1 Positive &$\mathcal{L}_{WAN}$  & 64.8 & 67.8 & 86.5 & 87.1 & \underline{46.3} & 47.5  \\ 
        (FSPL Setting)&$\mathcal{L}_{EPR}$ & 63.3 & 67.5 & 85.5 & 85.6 & 46.0 & 48.9  \\ 
        &$\mathcal{L}_{ROLE}$& 66.3 & 69.0 & \underline{87.9} & \underline{88.2} & 43.1 & \underline{51.0}  \\ 
      & \textbf{$\mathcal{L}_{PLMCL}$} & \textbf{71.1} & \textbf{72.0} & \textbf{89.0} & \textbf{88.7} & \textbf{48.6} & \textbf{51.5}  \\ \hline
    \end{tabular}
    \label{tab:fpsl}
\end{adjustbox}
\vskip-3.5mm
\end{table*}
\subsection{Full-set Single Positive Label (FSPL)}
This subsection compares different models under FSPL setting.
Table~\ref{tab:fpsl} reports the mAPs over three different datasets under two training settings: end-to-end and linear-init. Besides FSPL, we also include in Table~\ref{tab:fpsl} the mAPs of two models ($\mathcal{L}_{BCE}$ and $\mathcal{L}_{BCE-LS}$) based on full-set fully labeled (FFL) setting~\cite{cole2021multi}, as references to examine the performance of our model under FSPL setting.
Compared with the FFL setting using 100\% of the labels, the FSPL setting uses only around 1.2\% of the total number of labels. It is worth mentioning that the mAPs of our model $\mathcal{L}_{PLMCL}$ under FSPL remain at a comparable level. For instance, the mAP of our model on the NUS dataset is 48.6\%, which is only 4.0\% lower than the FFL setting $\mathcal{L}_{BCE}$.  Similar conclusions can be drawn in COCO and VOC datasets.  When comparing the approaches under FSPL setting (the bottom part of Table~\ref{tab:fpsl}), we observe that our model $\mathcal{L}_{PLMCL}$ always achieves the best performance on different datasets.

\noindent
\subsubsection{Backbone ResNet-101}.
Besides ResNet-50, we also conducted experiments using a different backbone, ResNet-101, in the end-to-end setting on two datasets under different SSPL settings.  The results are reported in Table~\ref{tab:resnet101}. Our model $\mathcal{L}_{PLMCL}$ achieves the hightest mAP scores under all settings. It is worth mentioning that our model ~$\mathcal{L}_{PLMCL}$ at 60\% of SSPL setting achieves higher mAP compared to~$\mathcal{L}_{ROLE}$ under FSPL setting (100\%) in COCO dataset, which demonstrates the efficiency of our model in saving labeling costs.
\begin{table*}[!t]
\caption{Quantitative results (mAP) of multi-label image classification on different subsets of the single observed label (SSPL) setting and FSPL (100\%) setting on
two different datasets with end-to-end learning using the Resnet-101 backbone. Bold represents the highest mAP and underline represents the second-best.}
\centering{
\begin{adjustbox}{width=0.95\textwidth} 
\begin{tabular}
    {
      |p{0.2\textwidth} |
      >{\centering}p{0.07\textwidth} |
      >{\centering}p{0.07\textwidth} |
      >{\centering}p{0.07\textwidth} |
      >{\centering}p{0.07\textwidth} |
      >{\centering}p{0.07\textwidth} ||
      >{\centering}p{0.07\textwidth} |
      >{\centering}p{0.07\textwidth} |
      >{\centering}p{0.07\textwidth} |
      >{\centering}p{0.07\textwidth} |
      >{\centering\arraybackslash}p{0.06\textwidth}|
    }
\hline
&\multicolumn{5}{c||}{\textbf{COCO Dataset}}&\multicolumn{5}{c|}{\textbf{VOC Dataset}}\\
\cline{2-11}
Models&\multicolumn{4}{c|}{SSPL}&\multicolumn{1}{c||}{FSPL}&\multicolumn{4}{c|}{SSPL}&\multicolumn{1}{c|}{FSPL}\\
\cline{2-11}
&20\%&40\%&60\%&80\%&100\%&20\%&40\%&60\%&80\%&100\%\\
\hline
$\mathcal{L}_{AN}$&42.0&52.4&57.9&62.3&63.9 &52.9&73.6&82.1&85.1&86.1\\
$\mathcal{L}_{AN-LS}$&49.2&57.5&61.8&65.6&66.9 &67.1&80.6&85.8&86.8&88.3\\
$\mathcal{L}_{WAN}$&\underline{57.3}&\underline{61.8}&\underline{64.9}&\underline{66.0}&66.7 &\underline{76.1}&\underline{83.1}&85.9&86.5&87.2\\
$\mathcal{L}_{EPR}$&51.8&57.6&60.9&62.8&63.8 &75.1&80.3&85.0&84.6&86.7\\
$\mathcal{L}_{ROLE}$&46.2&58.3&62.8&65.5&\underline{67.3}  &66.7&82.3&\underline{86.2}&\underline{87.4}&\underline{88.8}\\

\textbf{$\mathcal{L}_{PLMCL}$ (ours) }&\textbf{61.4}&\textbf{67.3}&\textbf{69.5}&\textbf{71.5}&\textbf{72.2} &\textbf{82.0}&\textbf{86.2}&\textbf{88.4}&\textbf{88.8}&\textbf{89.7}\\
\hline
\end{tabular}
\end{adjustbox}
}
\label{tab:resnet101}
\vskip-4mm
\end{table*}
\subsection{Ablation Study.} 
\label{sub:ablation}
\vskip-3.5mm
\noindent
\subsubsection{PLMCL Modules.} 
Our ablation analysis highlights the importance of each component in our method: the momentum $m_t$, the scheduler $\xi$, and the self-guided factor $\psi$. The results are reported in Table~\ref{tab:ablation1}. We start with the basic variation of the model which uses the cross-entropy loss on the observed labels.
\begin{wraptable}{r}{5.5cm}
\vskip-6mm
\caption{Ablation study for all proposed model components on SSPL (20\%) and FSPL with linear-init training setting on COCO dataset.}
\vskip-4mm
\label{tab:ablation1}
\centering{
\begin{adjustbox}{width=0.45\textwidth}  
\begin{tabular}
    {
      p{0.05\textwidth} 
      >{\centering}p{0.05\textwidth} 
      >{\centering}p{0.05\textwidth} |
      >{\centering}p{0.15\textwidth} |
      >{\centering\arraybackslash}p{0.08\textwidth}
    }
\hline
\multicolumn{3}{c|}{Hyper Parameters}&\multicolumn{2}{c}{COCO Dataset (mAP \%)}\\
\hline
 $m_t$&$\xi$&$\psi$& $20\%   $&100\%\\
\hline
&&&59.0&67.3\\
\checkmark &&&62.0&69.9\\
\checkmark &\checkmark&&63.1&70.4\\
\checkmark&\checkmark&\checkmark&64.8&72.0\\
\hline
\end{tabular}
\end{adjustbox}
}
\vskip-6mm
\end{wraptable}
 In this case, we assume that all the unobserved labels are negative, since ignoring the unobserved labels may lead to overfitting.
When adding components into the model, this assumption can be removed since we use the pseudo labels to deal with the unobserved labels.  We first add the momentum generation module to update the soft pseudo labels which improve the mAP for different settings and datasets (Table~\ref{tab:ablation1}, second row). Adding the confidence-aware scheduler and the self-guided factor further enhances in the overall performance, since they work together to make the training adaptive with respect to the confidence of the pseudo labels (Table \ref{tab:ablation1}, last two rows).   

\noindent
\subsubsection{Hyper Parameters.} 
The influence of the hyper-parameters is studied on the VOC validation set under the $40\%$ SSPL setting. We run experiments with different hyper-parameters.  In all of these experiments, our model remains stable and reliable. Table~\ref{tab:hyperparameters} shows the variations in mAP according to the changes of hyper-parameters $\beta_1$, $\beta_2$, and $\lambda$. Recall that the hyper-parameter $\beta_1$ is used to balance the previous network knowledge and the current decision of the network.
\begin{wraptable}{r}{6.5cm}
\centering
\vskip-1mm
    \caption{mAP score (\%) of our method as a function of  $\beta_1$, $\beta_2$, $\lambda$ in subset single positive setting SSPL at $40\%$ using end-to-end training setting on the
    VOC validation set.}
\vskip-4mm
    \begin{adjustbox}{width=0.53\textwidth} 
    \begin{tabular}
            {
     | p{0.39\textwidth} |
      >{\centering}p{0.06\textwidth} |
      >{\centering}p{0.06\textwidth} |
      >{\centering}p{0.06\textwidth} |
      >{\centering\arraybackslash}p{0.06\textwidth}|
    }
    \hline
         \textbf{$\beta_1$ } & 0.1& 0.5 & 0.7 & 0.9  \\ 
        \hline
        mAP (\%) ($\beta_2 = 0.6$)& 84.4 & 84.5 & \textbf{84.6} & 84.0   \\ \hline
                \hline
        $\beta_2$ & 0.1 & 0.4 & 0.6 & 0.9  \\ \hline
        mAP (\%) ($\beta_1=0.7$)&83.5 &84.5 & \textbf{84.6} & 84.5   \\ \hline
                \hline
        $\lambda$ &1 & 2 & 3 & 4 \\\hline
        mAP (\%) ($\beta_1=0.7$, $\beta_2 = 0.6$) &83.0 &83.2  &83.9  & \textbf{84.6}  \\ \hline
    \end{tabular}
    \end{adjustbox}
    \label{tab:hyperparameters}
\vskip-5mm
\end{wraptable}
We observed that the mAP score improves when the balance is achieved ($\beta_1$ is close to 0.5) and decreases in the extreme cases ($\beta_1 = 0.1$ emphasizing the current decision of the network and $\beta_1 = 0.9$  emphasizing the previous network knowledge). 
$\lambda$ demonstrates the effectiveness of the self-guided factor $\psi$ that controls the amount of momentum contributing to the update of pseudo levels.  The results in Table~\ref{tab:hyperparameters} indicates that the mAP score increases as the $\lambda$ increases.
When increasing the scheduler parameter $\beta_2$, the mAP score almost remains the same level.  Overall, our model is not very sensitive to the variations in these hyper-parameters. 

\subsubsection{Comparison with Semi-supervised Models.} 
\begin{wrapfigure}{r}{0.49\textwidth}
    	\vskip-4mm
  \centering
  	\includegraphics[scale=0.45]{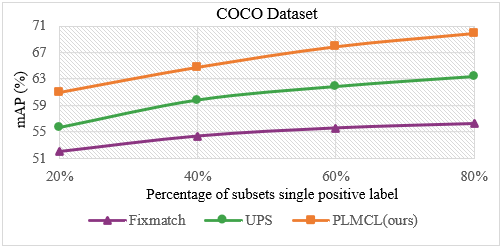}\\
  	\vskip-3mm
  \caption{Ablation study on the difference between using our method PLMCL and semi-supervised models. }
  \label{fig:ensembling}
    	\vskip-4mm
\end{wrapfigure}

We compare our PLMCL with the existing semi-supervised models such as FixMatch \cite{sohn2020fixmatch} and UPS~\cite{rizve2021in} on COCO dataset. Fig.~\ref{fig:ensembling} shows the comparison of the mAP results under different
SSPL settings. Notice that PLMCL always outperforms the other two methods.
This is because FixMatch and UPS are trained based on the supervised loss over the set of observed labels while ignoring the unobserved labels of the labeled set as explained in subsection \ref{semi-supervised}.
Therefore,  there is a lack of continuity in pseudo label updating. On the contrary, PLMCL can guarantee such continuity and capture the temporal information during training.  So it is not surprising that PLMCL achieves the best mAP score.

\noindent
\section{Conclusions}
\label{sec:con}
This paper presents a momentum-based curriculum learning method for multi-label image classification under partial-label settings. In PLMCL, soft pseudo labels are generated per image and then updated based on the momentum and the confidence levels. Through extensive experiments on three different partial-label settings, including our proposed new SSPL setting, we demonstrated that our PLMCL outperforms the state-of-the-art methods, especially when training on fewer observed labels. \\

\noindent
\textbf{Acknowledgements.}
The authors gratefully acknowledge the partial financial support of the National Science Foundation (1830512).
\clearpage
\bibliographystyle{splncs04}
\bibliography{eccv2022}
\end{document}